\documentclass[10pt,twocolumn,letterpaper]{article}

\usepackage{wacv}              

\usepackage{graphicx}
\usepackage{amsmath}
\usepackage{amssymb}
\usepackage{booktabs}
\usepackage{svg}
\usepackage[export]{adjustbox} 
\usepackage{arydshln}
\usepackage[accsupp]{axessibility}

\newcommand{\blue}[1]{\textcolor{black}{#1}}

\usepackage[pagebackref,breaklinks,colorlinks]{hyperref}

\usepackage[capitalize]{cleveref}
\crefname{section}{Sec.}{Secs.}
\Crefname{section}{Section}{Sections}
\Crefname{table}{Table}{Tables}
\crefname{table}{Tab.}{Tabs.}

\def\wacvPaperID{363} 
\def\confName{WACV}
\def\confYear{2025}

\begin{document}

\title{Inferring Past Human Actions in Homes with Abductive Reasoning}

\author{
Clement Tan$^{1}$ \quad Chai Kiat Yeo$^{1}$ \quad Cheston Tan$^{2}$ \quad Basura Fernando$^{1,2}$\\
$^{1}$Nanyang Technological University, Singapore\\
$^{2}$Centre for Frontier AI Research (CFAR), Agency for Science, Technology and Research (A*STAR) \\ 
{\tt\small \{s190099, asckyeo\}@ntu.edu.sg, \{cheston$-$tan@i2r, fernando\_basura@ihpc\}.a$-$star.edu.sg}
}

\maketitle

\begin{abstract}
Abductive reasoning aims to make the most likely inference for a given set of incomplete observations.
In this paper, we introduce ``Abductive Past Action Inference", a novel research task aimed at identifying the past actions performed by individuals within homes to reach specific states captured in a single image, using abductive inference.
The research explores three key abductive inference problems: past action set prediction, past action sequence prediction, and abductive past action verification.
We introduce several models tailored for abductive past action inference, including a relational graph neural network, a relational bilinear pooling model, and a relational transformer model. 
Notably, the newly proposed object-relational bilinear graph encoder-decoder (BiGED) model emerges as the most effective among all methods evaluated, demonstrating good proficiency in handling the intricacies of the Action Genome dataset. 
\blue{The contributions of this research significantly advance the ability of deep learning models to reason about current scene evidence and make highly plausible inferences about past human actions. This advancement enables a deeper understanding of events and behaviors, which can enhance decision-making and improve system capabilities across various real-world applications such as Human-Robot Interaction and Elderly Care and Health Monitoring.}
Code and data available at \url{https://github.com/LUNAProject22/AAR}
\end{abstract}

\section{Introduction}
\label{sec:intro}

Reasoning is an inherent part of human intelligence as it allows us to draw conclusions and construct explanations from existing knowledge when dealing with an uncertain scenario. One of the reasoning abilities that humans possess is abductive reasoning. Abductive reasoning aims to infer the most compelling explanation for a given set of observed facts based on a logical theory. 
\blue{In this work, we study the new problem of inferring past human actions from visual information using abductive inference.}
It is an extremely useful tool in our daily life, as we often rely on a set of facts to form the most probable conclusion.  
\blue{In fact, a comprehensive understanding of a situation requires considering both past and future information. The ability to perform abductive reasoning about past human actions is vital for human-robot collaboration AI-assisted accident and crime investigation and assistive robotics. Furthermore, robots working in dynamic environments benefit from understanding previous human actions to better anticipate future behaviour or adapt their actions accordingly.} 
Imagine the scenario where a rescue robot enters an elderly person's house to check on why he or she is not responding to the routine automated phone call. 
Upon entering the house, the robot observes its surroundings and notices that the \textit{back door is left open} but \textit{nothing else is out of the ordinary}. These observations may form a basis for a rational agent -- the elderly might have \textit{opened the door} and \textit{went into the garden}. The robot can immediately make its way to search for him/her in the back garden. 
This example illustrates how a social rescue robot can utilize observed facts from the scene to infer past human actions, thereby reasoning about the individual's whereabouts and ensuring their safety through abductive reasoning.

In recent years, there have been some great initiatives made in abductive reasoning for computer vision~\cite{liang2022visual, zhang2024rca,hessel2022abduction}. In particular, \cite{liang2022visual} generates the description of the hypothesis and the premises in the natural language given a video snapshot of events. Without the generation of a hypothesis description, these methods boil down to dense video captioning. A similar task is also presented in~\cite{hessel2022abduction} where given an image, the model must perform logical abduction to explain the observation in natural language. 
The use of natural language queries in these tasks presents challenges related to language understanding, making the abductive reasoning task more complicated.

\begin{figure}[t]
   \centering
   \includegraphics[width=0.99\linewidth]{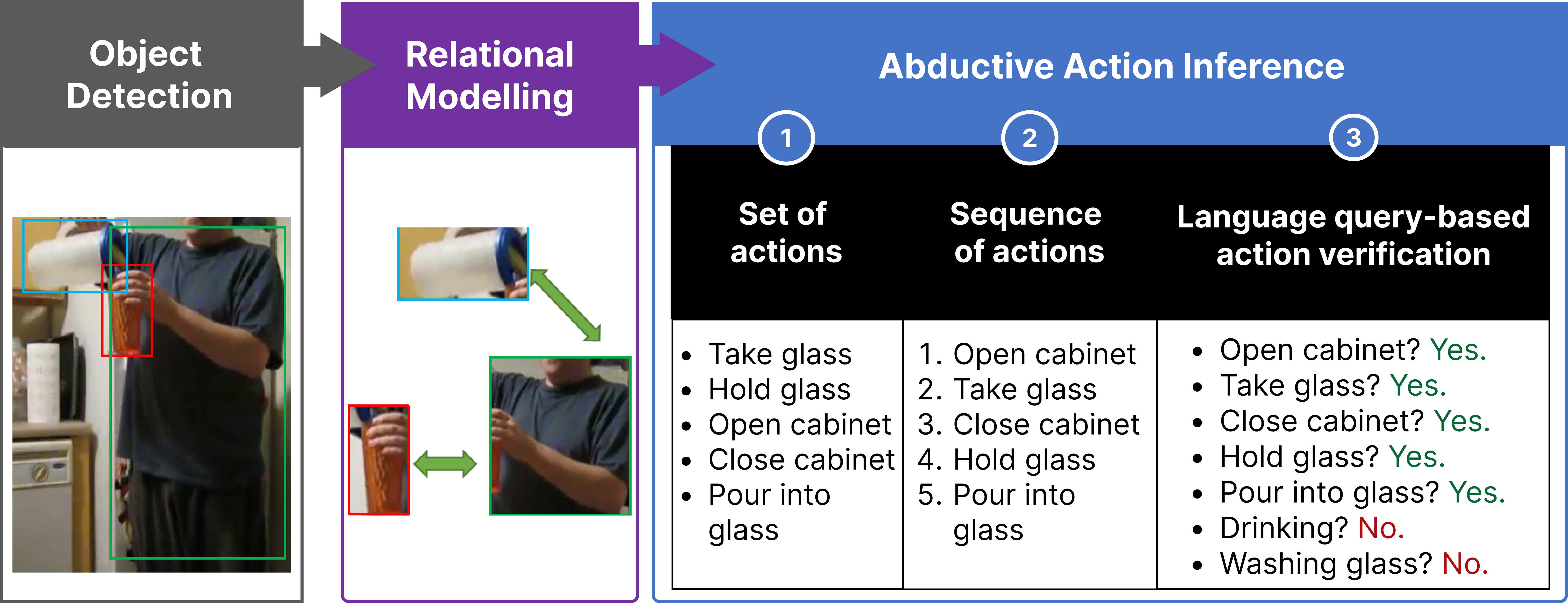}
   \caption{Proposed object-relational approach for abductive past action inference. Models are tasked to: 1) abduct the set of past actions, 2) abduct the sequence of past actions, and 3) perform abductive past action verification.}   
   \label{fig:my_label}
\end{figure}

In contrast to these recent works, we challenge a model to infer multiple human actions that may have occurred in the past from a given image. 
Based on the visual information from the image, objects such as a person, glass and cabinet, may provide clues from which humans can draw conclusions -- see Fig.~\ref{fig:my_label}. 
We term this new task, \emph{Abductive Past Action Inference} and further benchmark how deep learning models perform on this challenging new task. 
For this task, the models are not only required to decipher the effects of human actions resulting in different environment states but also solve causal chains of action effects, a task that can be challenging even for humans.
Furthermore, the task relies on the model's ability to perform abductive reasoning based on factual evidence i.e., determining what actions may have or have not been performed based on visual information in the image.
Humans can solve this task by using prior experience (knowledge) about actions and their effects and using reasoning to make logical inferences. Are deep learning models able to perform abductive past action inference by utilizing visual knowledge present in a given image and a learned understanding of the domain? We aim to answer this question in this paper.


Human action can be viewed as an evolution of human-object relationships over time. Therefore, the state of human-object relations in a scene may give away some cues about the actions that may have been executed by the human. 
We hypothesize that deep learning models are able to perform logical abduction on past actions using visual and semantic information of human-object relations in the scene. 
As these human-object relations provide substantial visual cues about actions and the effects of past actions, it makes it easier for the models to infer past actions.
On the other hand, there is also the duality in which the evidence should support those conclusions (the actions inferred by the model). If a human executed a set of actions $\mathcal{A}$ which resulted in a state whereby a human-object relation set $\mathcal{R}$ is formed as an effect of those executed actions (i.e., $\mathcal{A} \rightarrow \mathcal{R}$), then using the relational information, we can formulate the task by aiming to infer $\mathcal{A}$ from $\mathcal{R}$. 
Therefore, we argue that human-object relational representations are vital for abductive past action inference and provide further justifications in our experiments. 

In this work, our models rely on the human-centric object relation tuples such as (person, glass) and (person, closet) obtainable from a single image at the current point in time to perform abductive past action inference. 
One can see why these human-centric relations are vital for identifying past actions: the (person, glass) relation may lead to deriving actions such as (person-pouring-water, person-took-glass-from-somewhere) while (person, closet) may imply actions such as (person-opening-closet, person-closing-closet) see -- Figure \ref{fig:my_label}. 
Therefore, we use objects and their relationships in the scene to construct human-centric object relations within each image.  
These relations are made up of both visual and semantic features of recognized objects. To effectively model relational information, we use bilinear pooling and to model inter-relational reasoning, we use a new relational graph neural network (GNN). 
We propose a new model called BiGED that uses both bilinear pooling and GNNs to effectively reason over human-object relations and inter-relational information of an image to perform abductive inference on past actions effectively.
%

Our contributions are summarized as follows: 
(1) To the best of our knowledge, we are the first to propose the abductive past action inference task, which involves predicting past actions through abductive inference. 
%
(2) We benchmark several image, video, vision-language, and object-relational models on this problem, thereby illustrating the importance of human-object relational representations. Additionally, we develop a new relational rule-based inference model which serve as relevant baseline models for the abductive past action inference task.
(3) We propose a novel relational bilinear graph dncoder-decoder model (BiGED) to tackle this challenging reasoning problem and show the effectiveness of this new design.

\section{Related Work}
\label{sec:related}
Our work is different from action recognition~\cite{jhuang2013towards, herath2017going} in a fundamental way. First, in action recognition, the objective is to identify the actions executed in the visible data (e.g., a video or an image in still image action recognition~\cite{guo2014survey}). In action recognition, the models can learn from visual cues what the action looks like and what constitutes an action. 
In our work, we aim to infer past actions that the model has never seen the human performing. The model only sees visual evidence (e.g. human-object relations) in the scene which is the outcome of executed actions. 
\blue{There are no motion cues or visual patterns of actions that the model can rely on to predict past actions.}
From a single static image, the machine should infer what actions may have been executed. This is drastically different from classical action recognition and action anticipation tasks. 

Abductive past action inference shares some similarity to short-term action anticipation~\cite{furnari2019would,fernando2021anticipating} and long-term action anticipation~\cite{abu2018will}. 
However, there are several notable differences between the two tasks. 
Firstly, in abductive past action inference, the goal of the model is to identify the most plausible actions executed by a human in the past based on the current evidence, whereas, in action anticipation, the model learns to predict future action sequences from current observations.
The primary distinction lies in abductive past action inference, where observations (evidence) may imply certain past actions, contrasting with action anticipation tasks that predict future actions without certainty.
\blue{In other words, in abductive past action inference, the evidence and clues indicate possible actions executed in the past that resulted in the evidence or clues. However, in action anticipation, the clues~\cite{Roy_2024_WACV}, context~\cite{gong2022future}, current actions~\cite{Girdhar_2021_ICCV}, and knowledge about the task~\cite{zhao2023antgpt} are used to infer probable future actions, but there is no guarantee that the predicted actions will be executed by a human.}
For instance, observing \textit{a person cleaning a room with a broom} suggests prior actions such as \textit{picking up the broom from somewhere} must have happened among many others. 
Even if \textit{putting away the broom} is anticipated somewhere in the future, other actions such as \textit{holding the broom} and \textit{opening a window} are also possible. 
\blue{Therefore, while action anticipation addresses the uncertainty of future human behavior, abductive past action inference models can utilize scene evidence (such as objects in the scene) to infer the most likely past actions.}
Additionally, in abductive past action inference, the uncertainty arises from the fact that several different actions may have resulted in similar states $\mathcal{R}$.
In our task, models should comprehend the consequences of each executed action and engage in abductive reasoning to infer the most probable set or sequence of past actions.
\blue{Another key difference between action anticipation and abductive past action inference is that in action anticipation, predictions made at time $t$ can leverage all past observations. In contrast, abductive past action inference relies solely on present and future information, where new future observations can potentially alter the evidence about past actions, making the inference process more challenging.}

Visual Commonsense Reasoning (VCR) \cite{zellers2019recognition,wu2019connective} and causal video understanding~\cite{parmar2024causalchaos,parmar2024learning} are also related to our work. In VCR \cite{zellers2019recognition}, given an image, object regions, and a question, the model should answer the question regarding what is happening in the given frame. The model has to also provide justifications for the selected answer in relation to the question. Authors in \cite{park2020visualcomet} also studied a similar problem where a dynamic story underlying the input image is generated using commonsense reasoning. In particular, VisualCOMET \cite{park2020visualcomet} extends VCR and attempts to generate a set of textual descriptions of events at present, a set of commonsense inferences on events before, a set of commonsense inferences on events after, and a set of commonsense inferences on people’s intents at present. In this vein, given the complete visual commonsense graph representing an image, they propose two tasks; (1) generate the rest of the visual commonsense graph that is connected to the current event and (2) generate a complete set of commonsense inferences. In contrast, given an image without any other natural language queries, we recognize visual objects in the scene and how they are related to the human, then use the human-centric object relational representation to infer the most likely actions executed by the human.

Recently, there are machine learning models that can also perform logical reasoning~\cite{dai2019bridging,zhong2023chatabl,cai2021abductive,han2023abductive,jin2022complex}. 
Visual scene graph generation~\cite{tang2020unbiased} and spatial-temporal scene graph generation~\cite{cong2021spatial} are also related to our work.
Graph neural networks are also related to our work ~\cite{scarselli2008graph,yu2020representative,schlichtkrull2018modeling}.
Our work is also related to bilinear pooling methods such as~\cite{Gao_2016_CVPR,fukui2016multimodal}.

\section{Abductive Past Action Inference}
\label{sec:method}
\textbf{Task:}
Given a single image, models have to infer past actions executed by humans up to the current moment in time. 
We name this task Abductive Past Action Inference.
Let us denote a human action by $a_i \in A$ where $A$ is the set of all actions and $E_1, E_2, \cdots $ is a collection of evidence from the evidence set $\mathcal{E}$.
As the evidence is a result of actions, we can write the logical implication $\mathcal{A} \rightarrow \mathcal{E}$ where $\mathcal{A}$ is the set of actions executed by a human which resulted in a set of evidence $\mathcal{E}$.
Then, the task aims to derive 1) the set of past actions, 2) the sequence of past actions that resulted in the current evidence shown in the image, and 3) abductive past action verification.
The abductive past action verification is a binary task where the model is given a single image and is required to answer a yes or no to an action query (did the person execute action $a_{x}$ in the past?).


\subsection{Object-Relational Representation Approach}
Our primary hypothesis is that human-object relations are essential for abductive past action inference. 
Therefore, we propose a human-object relational approach for the task. 
In all three tasks, our general approach is as follows. We make use of detected humans and objects in the image and then generate a representation for human-centric object relations. 
Then, using these human-centric object relations, we summarize the visual image, and using neural models, we infer the most likely actions executed by the human. The overview of this approach is shown in Figure~\ref{fig:my_label}.
Next, we first discuss abductive past action set inference, followed by the details of abductive past action verification. 

\paragraph{Abductive past action set prediction.} 
Let us denote the object by $o \in O$, the predicate category by $p \in P$, and the human by $h$. The $j^{th}$ relation $R_j$ is a triplet of the form $\left<h,p,o\right>$. 
In the $i^{th}$ image, we observe $n$ number of relations $\mathcal{R}_i = \{ R_1, R_2, \cdots R_n\}$ where  $\mathcal{R}_i$ is the relation set present in the situation shown in that image. These relations constitute the evidence ($\mathcal{E}$).
The relation set $\mathcal{R}_i$ is an effect of a person executing an action set/sequence $\mathcal{A}_i = \{ a_1, \cdots a_K \}$.
Therefore, the following association holds.
\begin{equation}
     \mathcal{A}_i 
 \rightarrow \mathcal{R}_i 
 \label{eq1}
\end{equation}
However, we do not know which action caused which relation (evidence), as this information is not available.
The association reveals that there is a lack of specific knowledge about the exact effects of individual actions and when multiple actions have been executed, the resulting effect is a combined effect of all executed actions.
Consequently, the learning mechanism must uncover the probable cause-and-effect relationships concerning actions.
Therefore, given $\mathcal{R}$ we aim to perform abductive past action inference to infer the most likely set of actions executed by the human using neural network learning.

We learn this abduction using the deep neural network functions $\phi()$, and $\phi_c()$.
The relational model, $\phi$ takes relation set as input and outputs a summary of relational information as a vector $x_r$.
\begin{equation}
x_r = \phi(R_1,\cdots,R_n;\theta_{\phi})
\label{eq.model1}
\end{equation}
The parameters of the relational model are denoted by $\theta_{\phi}$.
The linear classifier ($\phi_{c}$) having the parameters $\theta_{{c}}$, takes relational information as a vector $x_r$ as input and returns the conditional probability of actions given the relational evidence as follows:
\begin{equation}
P(a_1,\cdots,a_K|R_1,\cdots,R_n ) = \phi_{c}(x_r; \theta_{{c}}) 
\label{eq.model}
\end{equation}

The training {and inference} sets comprise images and corresponding action set $\mathcal{A}_i$. From each image, we extract the relation set $\mathcal{R}_i$. Therefore, the dataset consists of $\mathcal{D} = \bigcup_{i} \{\mathcal{R}_i,\mathcal{A}_i\}$.
Given the training set ($\mathcal{D}$), we learn the model function in Equations~\ref{eq.model1} and \ref{eq.model} using backpropagation as follows:
\begin{equation}
    \theta_{\phi}^{*}, \theta_{{c}}^{*} = \texttt{argmin}_{\theta_{\phi}, \theta_{{c}}}\sum_i -log(P(\mathcal{A}_i|\mathcal{R}_i ))
\end{equation}
where $\theta_{\phi}^{*}, \theta_{{c}}^{*}$ are the optimal parameters. As this is a multi-label-multi-class classification problem, we utilize the max-margin multi-label loss \blue{from PyTorch where the margin is set to 1}, during training. 


\paragraph{Abductive past action verification.}
Abductive verification model $\phi_{ver}()$ takes the evidence $\mathcal{E}$ and the semantic representation of the past action (e.g. textual encoding of the action name) $y_{a}$ as inputs and outputs a binary classification score indicating if the evidence supports the action or not, i.e. $\phi_{ver}(\mathcal{E},y_{a}) \rightarrow [0,1]$.
Specifically, we encode the past action name using the CLIP \cite{radford2021learning} text encoder to obtain the textual encoding $y_{a}$ for action class $a$. 
Then, we concatenate $y_a$ with $x_r$ and utilize a two-layer MLP to perform binary classification to determine whether action $a$ was executed or not. We use the max-margin loss to train $\phi_{ver}()$.
\textit{Note that semantic embedding of action classes ($y_a$) is not a necessity here. For example, one might learn the class embeddings from scratch} removing the dependency on language or use one-hot-vectors.

\subsection{Relational Representation}
\label{sec:feat}
To obtain the relation representation, we extract features from the human and object regions of each image using a FasterRCNN~\cite{ren2015faster} model with a ResNet101 backbone~\cite{he2016deep}. 
Let us denote the human feature by $x_h$, the object feature by $x_o$, and the features extracted from taking the union region of both human and object features by $x_u$. As we do not know the predicate or the relationship label for the relation between $x_h$ and $x_o$, we use the concatenation of all three visual features $x_h$, $x_o$, and $x_u$ as the joint relational visual feature $x_v = [x_h, x_o, x_u ]$. 
Using FasterRCNN, we can also obtain the object and human categories. We use Glove~\cite{pennington2014glove} embedding to acquire a semantic representation of each human and object in the image. Let us denote the Glove embedding of the human by $y_h$ and the object by $y_o$. Then, the semantic representation of the relation is given by $y_s = [y_h, y_o]$.
Using both visual and semantic representations, we obtain a joint representation for each human-centric object relation in a given image. Therefore, the default relation representation for a relation $R = <h,p,o>$ is given by $ r = [x_v, y_s] $.
Note that we do not have access to the predicate class or any information about the predicate. 
Next, we present several neural and non-neural models that we developed in this paper that uses relational representations for the abductive past action inference task. 

The details of abductive past action sequence inference are provided in the supplementary materials (section 4.1).
Next, we present our graph neural network model to infer past actions based on relational information.




\subsection{GNNED: Relational Graph Neural Network}
\label{sec:GNNED}
\begin{figure}[t]
    \centering
    \includegraphics[width=0.99\linewidth]{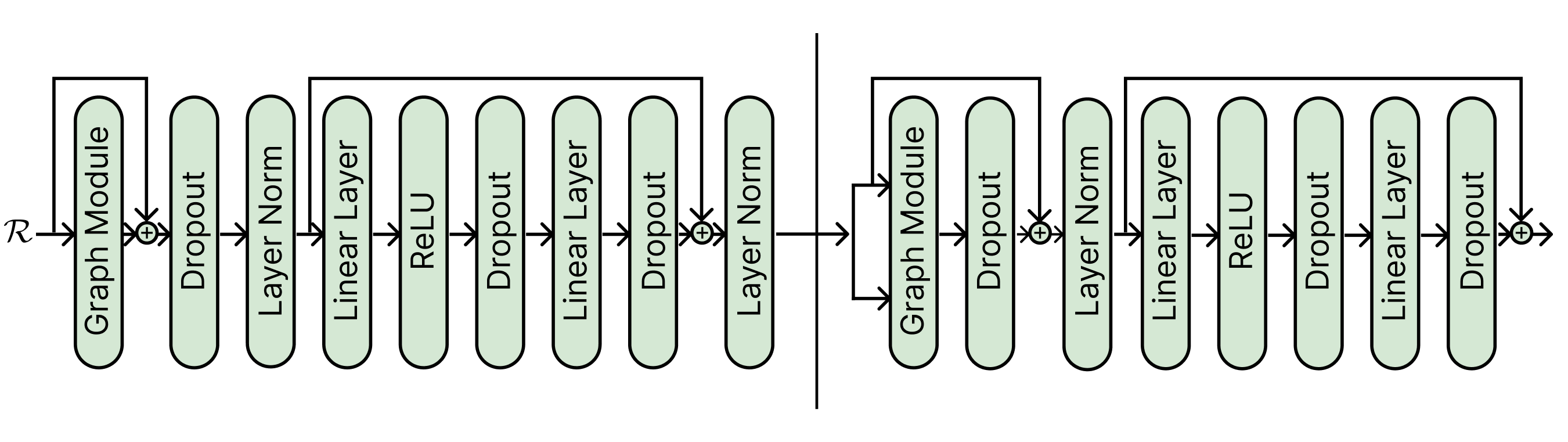}
    \caption{The graph neural network encoder (left) and graph neural network decoder (right) architecture. The residual connections are shown with the $+$ sign.}
    \label{fig:graph-encoder}
\end{figure}

The graph neural network-based encoder-decoder model summarizes relational information for abductive past action inference. Given the relational data with slight notation abuse, let us denote the relational representations by a $n\times d$ matrix $\mathcal{R} = [r_1, r_2, \cdots, r_n]$, where $r_n$ has $d$ dimensions. In our graph neural network encoder-decoder (GNNED) model, we first project the relational data using a linear function as follows:
\begin{equation}
   \mathcal{R'} = \mathcal{R}  W_l + b_l
   \label{eq.r}
\end{equation}
where $\mathcal{R'} = [r'_1, r'_2, \cdots, r'_n]$.
Then, we construct the affinity matrix $W_A(i,j)$  using Jaccard Vector similarity,
where $W_A(i,j)$ shows the similarity/affinity between the i-th relation and the j-th relation in the set.
Here, we use Jaccard Vector Similarity which is a smooth and fully differentiable affinity~\cite{fernando2021anticipating}. 
Note that Jaccard Vector Similarity is bounded by [-1,1]. 
Thereafter, we obtain the graph-encoded relational representation as follows:
\begin{equation}
   G_e = ReLU( (W_A   \mathcal{R'})  W_g + b_g)
   \label{eq.ge}
\end{equation}
where $W_g$ and $b_g$ are the weight matrix and bias term respectively. We call equations~\ref{eq.r}-\ref{eq.ge} the graph module. Using the graph module as a base model, we develop a graph encoding layer. 
The relational graph neural network encoder-decoder architecture we proposed is shown in Figure~\ref{fig:graph-encoder}.
Our graph encoder-decoder model consists of one graph encoder and three graph decoders by default.
The graph module is able to model the inter-relations between human-object relations and reasons about them to perform abduction on past actions. 
The graph encoding layer (left) is very similar to the Transformer encoder layer~\cite{vaswani2017attention}. 
The graph encoding layer consists of drop-out layers, layer norm~\cite{ba2016layer}, linear layers, and residual connections~\cite{he2016deep}.
The graph decoder layer (right) is also similar to a Transformer decoder layer except for the graph module.
Finally, we apply max-pooling at the end of the graph encoder-decoder model to obtain the final image representation $x_r$.
%
%
\begin{figure}[t]
    \centering
    \includegraphics[width=\columnwidth]{images/biged.png}
    \caption{The Bilinear Graph Encoder-Decoder (BiGED) architecture.}
    \label{fig:biGED}
\end{figure}
\subsection{RBP: Relational Bilinear Pooling}
{To effectively model the higher-order relational information between human and object features, we use bilinear pooling.}
Given the human representation $x_h$ and the object representation $x_o$, we use bilinear pooling with a weight matrix $W_b$ of size $d \times d \times d$ and linear projection matrices $W_{bl},W_{jb}$ as follows:
\begin{align}
    \label{eq.bo}
    o' &= ReLU(W_o x_o + b_o) \\ 
    \label{eq.bh}
    h' &= ReLU(W_h x_h  + b_h)\\ 
    \label{eq.br}
    r_b & =  ReLU([h' W_b o' ; ([h';o'] W_{bl} + b_{bl})])  W_{jb} + b_{jb} 
\end{align}
where [;] represents the vector concatenation and $h' W_b o'$ is the bilinear pooling operator applied over human and object features. 
$([h';o'] W_{bl} + b_{bl})$ is the output of concatenated human and object features followed by a linear projection using weight matrix $W_{bl}$ and bias term $b_{bl}$. 
{In contrast to bilinear pooling, the concatenated linear projection captures direct relational information between human and object features.}
Then, we concatenate the bilinear pooled vector ($h' W_b o'$) and the output of the linear projection ($([h';o'] W_{bl} + b_{bl})$). Next, we use ReLU and apply another linear projection ($W_{jb} + b_{jb}$).
Finally, we concatenate the overlap feature $x_u$ with the overall model output ($r_b$) and apply max-pooling across all relational features ($[r_b; x_u]$) in the image to obtain $x_r$.

\subsection{BiGED: Bilinear Graph Encoder-Decoder}
Finally, to take advantage of both bilinear relational modeling and graph neural network encoder-decoder models, we combine both strategies as shown in Fig~\ref{fig:biGED}. 
The main idea is to replace the projection function in Equation~\ref{eq.bo} with a graph neural network encoder-decoder model.
Let us denote the graph neural network encoder-decoder model by $f_{Ged}()$. Then, equation~\ref{eq.bo} will be replaced as follows:
\begin{align}
    \label{eq.go}
    O' &= f_{Ged}(X_o)     
\end{align}
where $X_o$ is all the object features in the image.
Afterward, we apply equation~\ref{eq.br} before using bilinear modeling to obtain the relational representation.
Note that as there are only one or two humans in the image, we do not use the GNNED to model human semantic features. 
The inputs to the BiGED model are the visual human features $x_h$, concatenated visual and semantic object features $[x_o,y_o]$ as well as the union features $x_u$.
Next, we concatenate the human and object features $[x_h,x_o,y_o]$ to obtain a joint feature and then pass it through a linear layer and another GNNED model.
The outputs of the bilinear, joint feature-based GNNED models {and overlap union feature $x_u$} are concatenated to obtain the final relational representation.
Afterward, we use max-pooling to obtain the representation $x_{r}$ for the image. For all models, we employ a linear classifier to infer past actions using the representation vector $x_{r}$. 


\section{Experiments and Results}
\label{sec:experiments}

\subsection{Action Genome Past Action Inference dataset}
We extend the Action Genome (AG) dataset \cite{ji2020action} and benchmark all models on the AG dataset for the abductive past action inference task.
Built upon the Charades dataset \cite{sigurdsson2016hollywood}, the AG dataset contains 9,848 videos with 476,000+ object bounding boxes and 1.72 million visual relationships annotated across 234,000+ frames.  
\textit{It should be noted that not all video frames in the Charades dataset are used in the AG dataset. Only a handful of keyframes are used in AG, and we follow the same.} 
The AG dataset does not provide action annotations. 
To obtain action annotations for images of AG, we leverage the Charades dataset which contains 157 action classes. 
The process of generating action sets and sequences using images from the Action Genome and action labels from the Charades dataset for the abductive past action inference task is detailed in Section 2 of the supplementary materials.

\subsection{Experimental Setup} 
After obtaining the action annotations of images for a given video, we drop videos having only one image as there are no past images and therefore, no past actions. 
For the remaining images, we assign action labels from the previous images in two different evaluation setups:

\noindent
\textbf{1. Abduct at \textbf{$T$}}: Given a image at time $T$, we add action labels from all the previous images to the ground truth (including actions from the current image) where $\mathcal{A}_{t}$ denotes all past actions of the $t^{th}$ image. Therefore, the ground truth action set $\mathcal{A}$ is given by $\mathcal{A} = \mathop{\cup}\limits_{t=1}^{T} \mathcal{A}_{t}$. 

\noindent
\textbf{2. Abduct last image}:
{Based on the first setup, we add an additional task where the model has to perform inference only on the last image of each video which contains all past actions.} If the last image is $T'$, then the action set is $\mathcal{A} = \mathop{\cup}\limits_{t=1}^{T'} \mathcal{A}_{t}$.
%
%
Note that in the Action Genome dataset, the images are sampled non-homogeneously from the Charades dataset videos. 
Therefore, the previous image occurs several seconds before the current image. 
In our abductive past action inference task, the ground truth past action sets are confined to the length of each video. 
We provide details on the number of images for a set of $n$ past actions in the AG dataset for these setups in the supplementary materials -- section 2 and figure 4. 

\subsection{Evaluation Metrics}\label{evaluation_metrics}
We utilize the mean Average Precision (mAP), {Recall@K (R@K), and mean Recall@K (mR@K)} metrics to evaluate the models for the abductive action set prediction and action verification tasks. 
Each image contains 8.9 and 8.2 actions for the \textbf{\textit{Abduct at $T$}} and \textbf{\textit{Abduct last image}} setups respectively. Therefore, K is set to 10 based on the average number of actions contained in a image. 
\blue{Please see the supplementary material Section 3 for more implementation details}. We will also release all our codes and models for future research.

\subsection{Baseline Models}
We benchmark several publicly available image (Resnet101-2D~\cite{he2016deep}, ViT~\cite{dosovitskiy2020image}) and video models (Slow-Fast~\cite{feichtenhofer2019slowfast} and Resnet50-3D) using the surrounding 8 frames of a image from the Charades dataset, \blue{and Video-Swin-S\cite{liu2022video}, Mvitv2~\cite{li2022mvitv2} and InternVideo~\cite{wang2022internvideo} models using future $K$ images from Action Genome to explore video based methods. The value of $K$ is set to the minimum possible frame size for each model, with the default being 5 frames}. 
Image models are pre-trained on ImageNet \cite{DBLP:journals/corr/RussakovskyDSKSMHKKBBF14} while video models are pre-trained on Kinetics400 \cite{kay2017kinetics} dataset and we fine-tune these models on our task.
We use a batch size of 32 with a learning rate of 1e-5. \blue{As for ViT, we use a batch size of 2042 using A100 GPUs}. 
\blue{All video-based methods are fine-tuned end-to-end.}
We also use CLIP linear-probe and zero-shot to perform abduction using several variants of the CLIP model~\cite{radford2021learning}.
The details of all other baseline models (Relational Rule-based inference, Relational MLP, and Relational Transformers) are presented in supplementary material Section 1.

\subsection{Human Performance Evaluation}
Human performance for the abductive past action set inference and verification tasks in the \textit{Abduct at $T$} setup is presented in Tables \ref{tab:main_results} and \ref{tab:main_results_4}. \blue{Performance on the abductive past action sequence inference is provided in the supplementary materials--see Table 1 and Section 4.1.}
All human experiments for the three sub-problems in the \textit{Abduct at $T$} setup follow the same procedure. Evaluators are asked to review 100 randomly sampled test images and manually assess all action classes in the Charades dataset without viewing the ground truth. They then select the likely past actions for each image. 

\subsection{Results: Abductive Past Action Set Prediction}
\begin{table}[t]
\centering
\resizebox{0.95\columnwidth}{!}{
\begin{tabular}{lccccc}
\hline
\textbf{Model} & \multicolumn{1}{l}{\textbf{mAP}} &  \textbf{R@10} & \textbf{mR@10} \\ \hline
Human Performance                 & -- & 80.60 & 82.81 \\ \hline
\multicolumn{4}{c}{\textbf{End-to-end training}} \\ \hline
ResNet101-2D \cite{he2016deep} & 9.27 & 18.63 & 11.51\\
ViT B/32 \cite{dosovitskiy2020image} & 7.27 & 16.84 & 8.82 \\ \hdashline
Resnet50-3D \cite{feichtenhofer2019slowfast} & 8.16 & 16.08 & 7.83 \\ 
Slow-Fast \cite{feichtenhofer2019slowfast}  & 7.91 & 14.42 & 7.65 \\ \hdashline
\blue{Video-Swin-S \cite{liu2022video}} - (\textcolor{red}{K=~5}) & \blue{14.86} & \blue{34.18} & \blue{19.05}\\ 
\blue{MvitV2 \cite{li2022mvitv2}} - (\textcolor{red}{K=16}) & \blue{14.01} & \blue{34.38} & \blue{15.17}\\ 
\blue{InternVideo \cite{wang2022internvideo}} - (\textcolor{red}{K=8}) & \blue{12.29} & \blue{30.72} & \blue{12.37}\\ 
\hline

\multicolumn{4}{c}{\textbf{Vision-language models}} \\ \hline
CLIP-ViT-B/32 (zero-shot) \cite{radford2021learning}    & 14.07 & 14.88 & 20.88 \\ 
CLIP-ViT-L/14 (zero-shot) \cite{radford2021learning}    & 19.79 & 21.88 & 27.77 \\ 
CLIP-ViT-B/32 (linear probe) \cite{radford2021learning} & 16.16  & 31.25   &  16.38\\ 
CLIP-ViT-L/14 (linear probe) \cite{radford2021learning} & 22.06 & 40.18 & 20.01 \\ 
\hline
\multicolumn{4}{c}{\textbf{Object-relational methods - using GT human/objects}} \\ \hline
Relational Rule-based inference & 26.27  & 48.94 & 36.89  \\
Relational MLP                  & 27.73\blue{$\pm$0.20}  & 42.50\blue{$\pm$0.68} & 25.80\blue{$\pm$0.61}\\
Relational Self Att. Transformer  & 33.59\blue{$\pm$0.17}  & 56.03\blue{$\pm$0.40} & 40.04\blue{$\pm$1.15} \\
Relational Cross Att. Transformer    & 34.73\blue{$\pm$0.05} & 56.89\blue{$\pm$0.47} & 40.75\blue{$\pm$0.57} \\
Relational GNNED                & 34.38\blue{$\pm$0.36}  & 57.17\blue{$\pm$0.35} & 42.83\blue{$\pm$0.21} \\
Relational Bilinear Pooling (RBP)  & 35.55\blue{$\pm$0.30}  & 59.98\blue{$\pm$0.68} & 43.53\blue{$\pm$0.63}   \\
BiGED                & \textbf{35.75\blue{$\pm$0.15}} & \textbf{60.55\blue{$\pm$0.41}} & \textbf{44.37\blue{$\pm$0.21}}    \\  \hdashline

BiGED - (\textcolor{red}{K=3})                & \blue{36.00 $\pm$ 0.12} & \blue{60.17$\pm$0.44} & \blue{42.82$ \pm$ 0.90} \\ 
BiGED - (\textcolor{red}{K=5})                & \blue{\textbf{37.34 $\pm$ 0.21}} & \blue{\textbf{61.16$ \pm$ 0.56}} & \blue{44.07 $\pm$ 0.87} \\ 
BiGED - (\textcolor{red}{K=7})                & \blue{36.57 $\pm$ 0.38} & \blue{ 60.65 $\pm$ 0.52} & \blue{43.12 $\pm$ 0.47} \\ 
\hline

\multicolumn{4}{c}{\textbf{Object-relational method - using FasterRCNN labels}} \\ \hline
BiGED                & \textbf{24.13\blue{ $\pm$ 0.04}} & \textbf{43.59\blue{ $\pm$ 0.88}} & \textbf{30.12\blue{ $\pm$ 0.23}}    \\ \hline
\end{tabular}
}
\caption{Abductive past action set inference performance using the proposed methods on the \textit{Abduct at $T$} setup.}
\label{tab:main_results}
\end{table}
\begin{table}[t]
\centering
\resizebox{0.95\columnwidth}{!}{
\begin{tabular}{lccccc}
\hline
\textbf{Model} & \multicolumn{1}{l}{\textbf{mAP}} &  \textbf{R@10} & \textbf{mR@10} \\ \hline
Relational Rule-based inference & 26.18 & 44.34 & 33.94 \\
Relational MLP                  & 25.99\blue{$\pm$0.10}  & 38.79\blue{$\pm$0.86} & 23.54\blue{$\pm$0.72} \\
Relational Self Att. Transformer  & 30.13\blue{$\pm$0.11}  & 47.55\blue{$\pm$0.14} & 35.05\blue{$\pm$0.55} \\
Relational Cross Att. Transformer & 31.07\blue{$\pm$0.20}  & 48.33\blue{$\pm$0.15} & 35.32\blue{$\pm$0.50}  \\
Relational GNNED                & 30.95\blue{$\pm$0.30}  & 48.18\blue{$\pm$0.17} & 36.36\blue{$\pm$0.12} \\
RBP                  & 31.48\blue{$\pm$0.20} & \textbf{49.79\blue{$\pm$0.55}} & \textbf{36.96\blue{$\pm$0.36}}\\
BiGED                & \textbf{31.41\blue{$\pm$0.15}} & 49.62\blue{$\pm$0.64} & 36.15\blue{$\pm$0.61} \\ \hline
\multicolumn{4}{c}{\textbf{Object-relational method - using FasterRCNN labels}} \\ \hline
BiGED                & \textbf{22.01\blue{$\pm$0.26}} & \textbf{37.06\blue{$\pm$0.52}} & \textbf{25.01\blue{$\pm$0.37}}    \\ \hline
\end{tabular}
}
\caption{Abductive past action set inference performance using the proposed methods on the \textit{Abduct last image} setup.}
\label{tab:main_results_2}
\end{table}

Our results for the abductive past action inference set prediction task are shown in Table \ref{tab:main_results}. 
These results are obtained based on the \textit{Abduct at $T$} setup.
During training, the model learns from every single image in the video sequence independently. Likewise, during inference, the model predicts the past action set on every single image.
The end-to-end trained models such as Slow-Fast~\cite{feichtenhofer2019slowfast}, ResNet50-3D, Resnet101-2D, and ViT perform poorly as it may be harder for these models to find clues that are needed for the abductive inference task. As there are no direct visual cues to infer previous actions (unlike object recognition or action recognition) from a given image, end-to-end learning becomes harder or near impossible for these models.
\blue{The Video-Swin-S Transformer model~\cite{liu2022video} shows promise in end-to-end models due to its use of future context (K future snapshots) and strong video representation capabilities.}

On the other hand, multi-modal foundational models such as the CLIP~\cite{radford2021learning} variants are able to obtain better results than vanilla CNN models on this task perhaps due to the quality of the visual representation.
Interestingly, object-relational models such as MLP and rule-based inference obtain decent performance.
One might argue that the performance of human-object relational models is attributed to the use of ground truth object labels in the scene. 
\blue{However, when we tried to incorporate ground truth objects using object bounding boxes with red colored boxes as visual prompts in the CLIP~\cite{radford2021learning} model, the performance was poor.}
The poor performance of the CLIP might be attributed to their training approach, which aims to align overall image features with corresponding text features. During their training, CLIP  assumes that the text in the captions accurately describes the visual content of the image. However, when it comes to abductive past action inference, no explicit visual cues are available to indicate the execution of certain actions.
We also note that the CLIP model demonstrates reasonable zero-shot performance. This may be because the CLIP model learns better vision features.

We experimented with generative models like ViLA~\cite{lin2023vila} (instruction tuning) and BLIP (question answering). \blue{Details are in the supplementary materials sections 1.4 for GPT-3.5 and section 1.5 for ViLA}. After instruction tuning on our dataset, ViLA achieved 10.5 mAP, 29.3 R@10, and 19.8 mR@10. We also tested GPT-3.5 with human-object relations as context, yielding 9.98 mAP, 25.22 R@10, and 20.32 mR@10. Due to the challenges of unconstrained text generation, models like BLIP, ViLA, and GPT-3.5 are excluded from the main comparison table.


The results also suggest that the human-object relational representations provide valuable evidence (cues) about what actions may have been executed in contrast to holistic vision representations.
Among object-relational models, the MLP model and rule-based inference perform the worst across all three metrics. 
The rule-based inference does not use any parametric learning and therefore it can only rely on \blue{statistics}. 
Interestingly, the rule-based method obtains similar performance to the MLP model indicating the MLP model merely learns from the bias of the dataset.

The relational transformer model improves results over MLP. 
Furthermore, the relational GNNED performance is comparable to the relational transformers. 
The transformer variants and GNNED have similar architectural properties and have better relational modeling capacity than the MLP model.
These models exploit the interrelation between visual and semantic relational representations to better understand the visual scene. This potentially helps to boost the performance of abductive past action inference.

Surprisingly, Relational Bilinear Pooling (RBP) obtains particularly good results outperforming the transformer and GNNED models.
The way relational reasoning is performed in RBP is fundamentally different from transformers and GNNED. 
The RBP models interactions between the human and object features within a relation more explicitly than the GNNED and Transformer.
However, unlike the GNNED or Transformer, RBP is unable to model interactions between relations.
Finally, the combination of GNNED and RBP, i.e., BiGED performs even better. 
This result is not surprising as BiGED takes advantage of better inter and intra-relation modelling.
\blue{We also experimented with a BiGED model, that takes $K$ future frames starting from frame at $T$ (i.e. Action Genome frames from $T$ to $T+K$) as inputs. The results of this experiment suggests that use of future snapshots helps improve performance.}

All object-relational models utilize the ground truth object labels from the AG dataset to obtain semantic representations. We observe a drop in performance when we use predicted objects from the FasterRCNN model. 
\blue{
Additionally, FasterRCNN-based object-semantic prediction performs worse than the visual-only BiGED model (\Cref{tab:semantic.visual.ablate}), indicating that incorrect semantics significantly harm performance.}
Nevertheless, the performance of BiGED with FasterRCNN labels is significantly better than end-to-end trained models and vision-language models.
Finally, it should be emphasized that human performance on this task is significantly better than any of the modern AI models, highlighting a substantial research gap in developing AI systems capable of effectively performing abductive past action inference.


\subsection{Results: Abduction on the Last Image}
\label{seq.exp.last}
We evaluated object-relational models on the second setup, where we perform abduction on the last image of each video, using models trained in the previous setup. Due to the variety of possible actions in a video sequence, this setup is more challenging. 
\blue{It should be noted that this is a special case of \textit{abduct at T}. This additional experiment allows us to observe and select the longest time horizon to determine whether the models are still able to abduct actions.}
Results in Table \ref{tab:main_results_2} show lower performance across all models compared to the previous setup, indicating the task's increased difficulty. The MLP model and rule-based inference perform poorly. The GNNED, RBP, and BiGED methods outperform the Transformer model, despite GNNED's similar architecture to the Transformer. BiGED achieves the highest mAP, while RBP excels in R@10 and mR@10.
\begin{table}[t]
\centering
\resizebox{0.95\columnwidth}{!}{
\begin{tabular}{lccccc}
\hline
\textbf{Model} & \multicolumn{1}{l}{\textbf{mAP}} &  \textbf{R@10} & \textbf{mR@10} \\ \hline
Human Performance               & -- &  92.26 & 93.71 \\
Relational MLP                  & 26.58\blue{$\pm$ 0.37}  & 41.71\blue{$\pm$ 0.82}  & 25.40\blue{$\pm$ 0.66} \\
Relational Self Att. Transformer          & 27.94\blue{$\pm$ 0.35}  & 45.72\blue{$\pm$ 1.42} & 30.12\blue{$\pm$ 2.30} \\
RBP                  & 32.19\blue{$\pm$ 0.44} & 53.76\blue{$\pm$ 0.89}  & 38.44\blue{$\pm$ 0.67}\\
BiGED                & \textbf{34.13\blue{$\pm$} \blue{0.39}} & \textbf{57.39\blue{$\pm$} \blue{0.10}}  & \textbf{41.97\blue{$\pm$} \blue{0.36}} \\ \hline
\end{tabular}
}
\caption{Abductive past action verification performance using the proposed methods on the \textit{Abduct at $T$} setup.}
\label{tab:main_results_4}
\end{table}


\subsection{Results: Abductive Past Action Verification}
We present abductive past action verification results in Table~\ref{tab:main_results_4} using the object-relational approach. We use the ground truth human and object class names to obtain the semantic representation. As the query is in textual form (i.e. the action class name), we suggest that the abductive past action verification resembles a human-like task. It is easy to answer yes, or no to the question ``Did the person execute action $a_i$ in this image to arrive at this state?"
Interestingly, the performance of this task is slightly lower than the main results we obtained in Table~\ref{tab:main_results}. Even though this task is mentally more straightforward for the human, it seems the task is slightly difficult for the machine as it now has to understand the complexities of human languages.

\subsection{\blue{Ablation on semantic vs visual}}
\blue{We use both visual and semantic features (Glove embedding of object names) to obtain the relational features --see~\Cref{sec:feat}. We ablate the impact of visual and semantic features on each model on Abductive Past Action Set Prediction (\textit{abduct at T}) and the results are shown in~\Cref{tab:semantic.visual.ablate}. 
While RBP achieves the best performance using ground truth object semantics, BiGED is the best-performing model for visual data alone by a considerable margin, making it the overall best method.
We can conclude while semantic features are effective, both visual and semantic features are complementary.}
\begin{table}[t]
    \scriptsize
    \centering
    {\begin{tabular}{lccc}\hline
         \blue{Model} & \blue{Visual Only} & \blue{Semantic Only} & \blue{Both} \\ \hline
         Rule-based inference & -- & 26.27 & -- \\
         MLP & 17.82 &	18.55 &	27.73 \\
         Transformer & 21.30 &	32.81 &	33.59 \\
         Relational GNNED & 21.55 &	32.82 & 34.38 \\
         RBP & 22.15 &	\textbf{33.03}	& 35.55 \\
         BiGED & \textbf{24.62} & 30.25 & \textbf{35.75} \\\hline
    \end{tabular}}
    \caption{\blue{mAP on Abductive Past Action Set Prediction (\textit{Abduct at T}) using visual and semantic features.}}
    \label{tab:semantic.visual.ablate}
\end{table}
\noindent
\blue{\textbf{Qualitative Results.} The qualitative results in supplementary material Section 4.4 demonstrate RBP and BiGED infer past actions more accurately.}
\blue{\textbf{Generalizability of BiGED.} A visual-only BiGED model trained to infer past actions was evaluated for action recognition at the video level. We obtained 50.5 mAP on the Charades dataset without any tuning. Although not state-of-the-art, these results suggest the value of the abductive past action inference model for general action understanding using only visual features.}

\section{Discussion \& Conclusion}
\label{sec:conclusion}

This paper introduces abductive past action inference, a task involving past action set prediction, sequence prediction, and verification, all formulated as closed-set classification tasks. Our experiments show that while deep learning models can perform these tasks to some extent, holistic end-to-end models are ineffective. Large multi-modal models like CLIP show promise, but our proposed human-object relational approaches—such as relational graph neural networks, bilinear pooling, and the BiGED model—outperform them, demonstrating the value of object-relational modelling. We find conditional text generation unsuitable for this task due to limited control, and even advanced foundational models fail after instruction tuning. Overall, human-object-centric video representations emerge as the most effective approach, and abductive past action inference may enhance general human action understanding.

\noindent
\textbf{Acknowledgment.} This research/project is supported by the National Research Foundation, Singapore, under its NRF Fellowship (Award NRF-NRFF14-2022-0001) and this research is partially supported by MOE grant RG100/23. It was also partly funded by an ASTAR CRF award to Cheston Tan and supported by the ASTAR CIS (ACIS) Scholarship awarded to Clement Tan. Any opinions, findings and conclusions or recommendations expressed in this material are those of the author(s) and do not reflect the views of these organizations.


{\small
\bibliographystyle{ieee_fullname}
\bibliography{main}
}

\clearpage
\section{Supplementary - Other Baseline Methods}
In this section, we give the details of all other baseline methods used in the paper.

\subsection{Rule-based Abductive Past Action Inference}
In abductive past action inference, we assume the following logical association holds,
\begin{equation}
     \{a_1, a_2, a_3, \cdots, a_K \}
 \rightarrow \{R_1,  R_2, \cdots, R_N \}
\end{equation}
where $\{R_1,  R_2, \cdots, R_N \}$ is the relation set $\mathcal{R}$ present in an image and $\{a_1, a_2, a_3, \cdots, a_K \}$ is the action set $\mathcal{A}$ executed by the human to arrive at the image. 
Note the set of all actions is denoted by $A$ where $\mathcal{A} \subset A$.

In rule-based inference, each relation is in the symbolic form $R_k = <H, o_k>$ where  $H$ and $o_k$ are the human feature and $k^{th}$ object label in the image. As the human feature is common in all relations, we omit the human feature in each relation. Then, the relational association is updated as follows:
\begin{equation}
     \{a_1, a_2, a_3, \cdots, a_K \} 
 \rightarrow \{o_1, o_2, \cdots, o_N\} 
\end{equation}
for any image.
In rule-based abductive past action set inference, for each given object pattern $\{o_1, o_2, \cdots, o_N\}$, we count the occurrence of each action $a_j$. Let us denote the frequency of action $a_j$ for object pattern $\mathcal{O}_q = \{o_1,  o_2, \cdots,  o_N\}$ from the entire training set by $C_j^q$. Therefore, for each object pattern $\mathcal{O}_q$, we obtain a frequency vector over all past actions denoted by:
\begin{equation}
     \mathbf{C^q} = [C_1^q, C_2^q, \cdots, C_{|A|}^q]
\end{equation}
Then, we can convert these frequencies into probabilities using softmax:
\begin{equation}
     P(A|\mathcal{O}_q) = softmax([C_1^q, C_2^q, \cdots, C_{|A|}^q])
     \label{eq}
\end{equation}
We use this to perform abductive past action set inference using the test set. Given a test image, we first obtain the object pattern $\mathcal{O}$. Next, we obtain the action probability vector for the object pattern from the training set using Equation~\ref{eq}. If an object pattern does not exist in the training set, we assign equal probability to each action.


\subsection{Relational MLP}
The MLP consists of 2-layers. The human feature $x_{h}$, object feature $x_{o}$, and union region of both human and feature $x_{u}$ obtained from the ResNet-101 FasterRCNN backbone are concatenated to form the joint relational visual features $x_{v}$. The semantic representation $y_{s}$ is formed via a concatenation of the Glove~\cite{pennington2014glove} embedding of the human $y_{h}$ and object $y_{o}$. We perform max pooling on the relational features, $\mathcal{R}_{i} = r_{1},r_{2},...,r_{n}$ in a given image, where each $r_{n} = [x_{v}, y_{s}]$ is the concatenation of visual and semantic features. Afterward, we pass these features into the 2-layer MLP. The inputs and outputs of the first layer are D-dimensional and we apply dropout with $p$ = 0.5. The last layer of the MLP is the classification layer. Lastly, we apply a sigmoid function before applying multi-label margin loss to train the model.

\subsection{Relational Transformer}
Transformers~\cite{vaswani2017attention} are a popular class of models in deep learning. They are effective at capturing relationships between far-apart elements in a set or a sequence. In this work, we use transformers as a set summarization model.\\
\noindent
\emph{Multi-head self-attention Transformer:}
Specifically, we utilize a multi-head self-attention (MSHA) transformer model. The MHSA transformer contains one encoder and three decoder layers by default.
We do not use any positional encoding as we are summarising a set.
Given the set of relational representation of an image $\mathcal{R}_i = r_1, r_2, \cdots, r_n$, the transformer model outputs a Tensor of size $n \times d$ where $d$ is the size of the relational representation.
Afterward, we use max-pooling to obtain an image representation vector $x_r$. A visual illustration of this model is shown in Fig.~\ref{fig:sat} (left).\\
\noindent
\emph{Cross-attention Transformer:}
Similar to the multi-head self-attention transformer, we use one encoder and three decoder layers. The inputs to the transformer encoder comprise concatenated visual and semantic features of a human and objects $[x_{h}, y_{h}, x_{o}, y_{o}]$, excluding the union features $x_{u}$.

\begin{figure}[t]
    \centering
    \includegraphics[width=0.49\columnwidth]{images/sat.png}
    \includegraphics[width=0.49\columnwidth]{images/cat.png}
    \caption{(left) The relational multi-head self-attention transformer. (right) The relational cross-attention transformer.}
    \label{fig:sat}
\end{figure}

\subsection{Relational GPT-3.5 Past Action Inference}
\begin{figure}[t]
    \centering
    \includegraphics[width=\columnwidth]{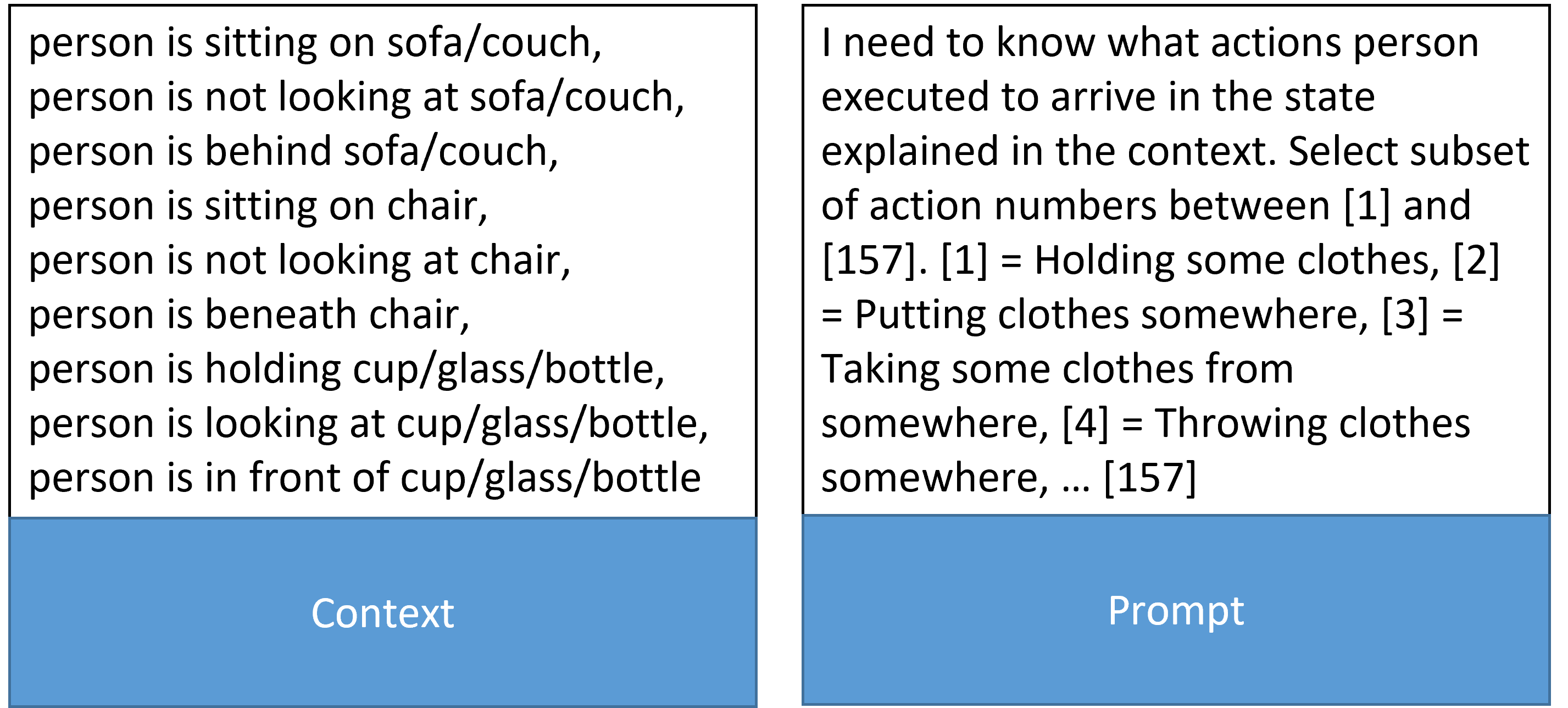}
    \caption{The context description and the textual prompt used for the GPT-3.5 turbo model.}
    \label{fig.chatgpt}
\end{figure}

\begin{figure}[t]
    \centering
    \includegraphics[width=\columnwidth]{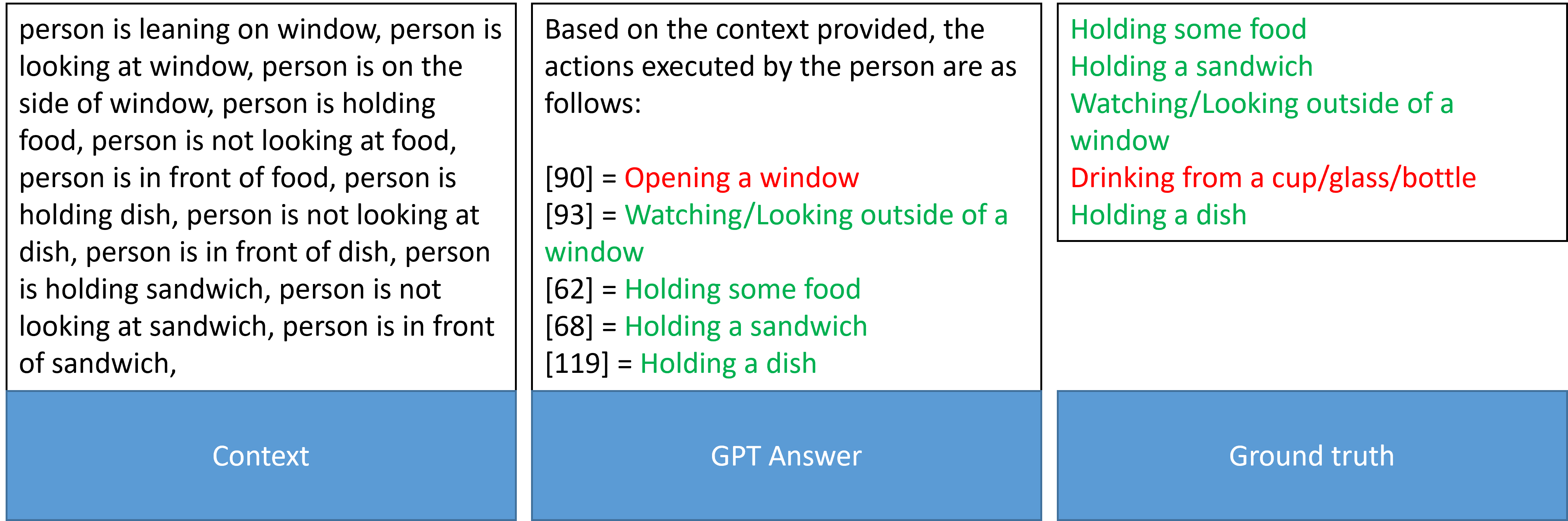}
    \caption{Answer generated by GPT-3.5 turbo model. The correct answers are shown in green color whereas false positives and negatives are shown in red. This example is cherry-picked.}
    \label{fig.chatgpt_ans}
\end{figure}

GPT and later versions~\cite{radford2018improving, radford2019language,brown2020language} have revolutionized the AI field by solving many natural language processing and reasoning tasks.
Here, we use the GPT-3.5 turbo version to perform abductive past action inference. To do this, we generate a query prompt as well as a contextual description for each image using the ground truth relational annotations based on the subject-predicate-object triplet relation. 
In contrast to the all other methods, we utilize the ground truth predicate label for GPT-3.5.
An example of the contextual description and textual prompt is shown in Figure~\ref{fig.chatgpt}. 
In addition, an answer generated by GPT-3.5 is shown in Figure~\ref{fig.chatgpt_ans}.
We specifically created the prompt such that GPT-3.5 responses are constrained to the ground truth action sets within the dataset.
Based on the responses from the GPT-3.5 model, we construct the score vector where the predicted action is marked with a score of 1 or 0 otherwise. We call this hard matching as we add 1 if and only if the GPT-3.5 model outputs the exact action class name given in the input prompt.

The GPT-3.5 model is able to generate reasonable answers in some images (see Fig~\ref{fig.chatgpt_ans}). However, most of the time GPT-3.5 answers are either overly conservative or aggressive. For example, GPT responds \emph{``There is not enough information given in the context to determine the specific actions the person executed to arrive in the described state"} and in some instances, it selects all action classes.
This may be the main reason for the poor performance of GPT-3.5. However, it should be noted that the GPT model is fed with more information than all other baselines as we also provide the predicate relation to the GPT-3.5 model. We also note that the GPT-3.5 + CLIP (Text) model with both soft and hard scores performs better than the hard score method.
Assuming that large language models such as GPT-3.5 are capable of human-like reasoning, we can perhaps suggest that abductive inference requires more than text-based reasoning and commonsense reasoning. Given the fact that pure rule-based inference performs better than GPT-3.5 with lesser information may suggest that GPT-3.5 is not suited for abductive past action inference due to it not having a detailed understanding of some of the human behaviors and effects of human actions.

\subsection{VILA Fine-tuning for Past Action Infernce}
With the proven success of Large Language Models (LLMs) across various NLP tasks, recent research has extended their capabilities towards vision tasks, resulting in the development of Visual Language Models (VLMs). These models are typically enhanced through prompt-tuning (where LLMs are frozen) or fine-tuning methods. We employ a fine-tuned VLM, VILA~\cite{lin2023vila}, which has not only advanced state-of-the-art performance in vision tasks but also retains robust capabilities in text processing. VILA demonstrates strong reasoning abilities in multi-image analysis, contextual learning, and zero/few-shot learning scenarios. Hence, we leverage VILA for the task of abductive past action set inference.

\section{Details on Dataset Creation}

\begin{figure}
    \centering
    \includegraphics[width=0.45\linewidth]{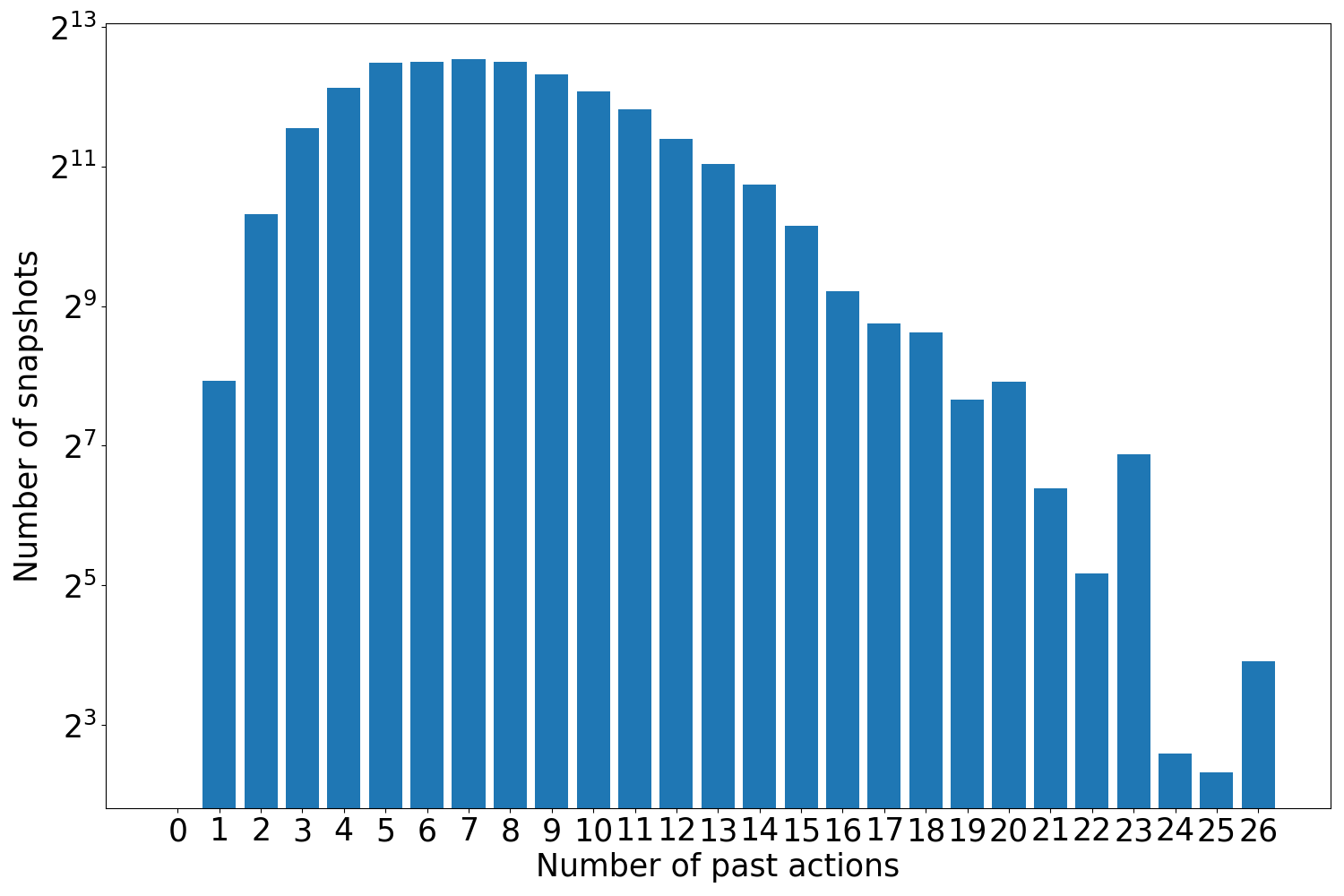}
    \includegraphics[width=0.45\linewidth]{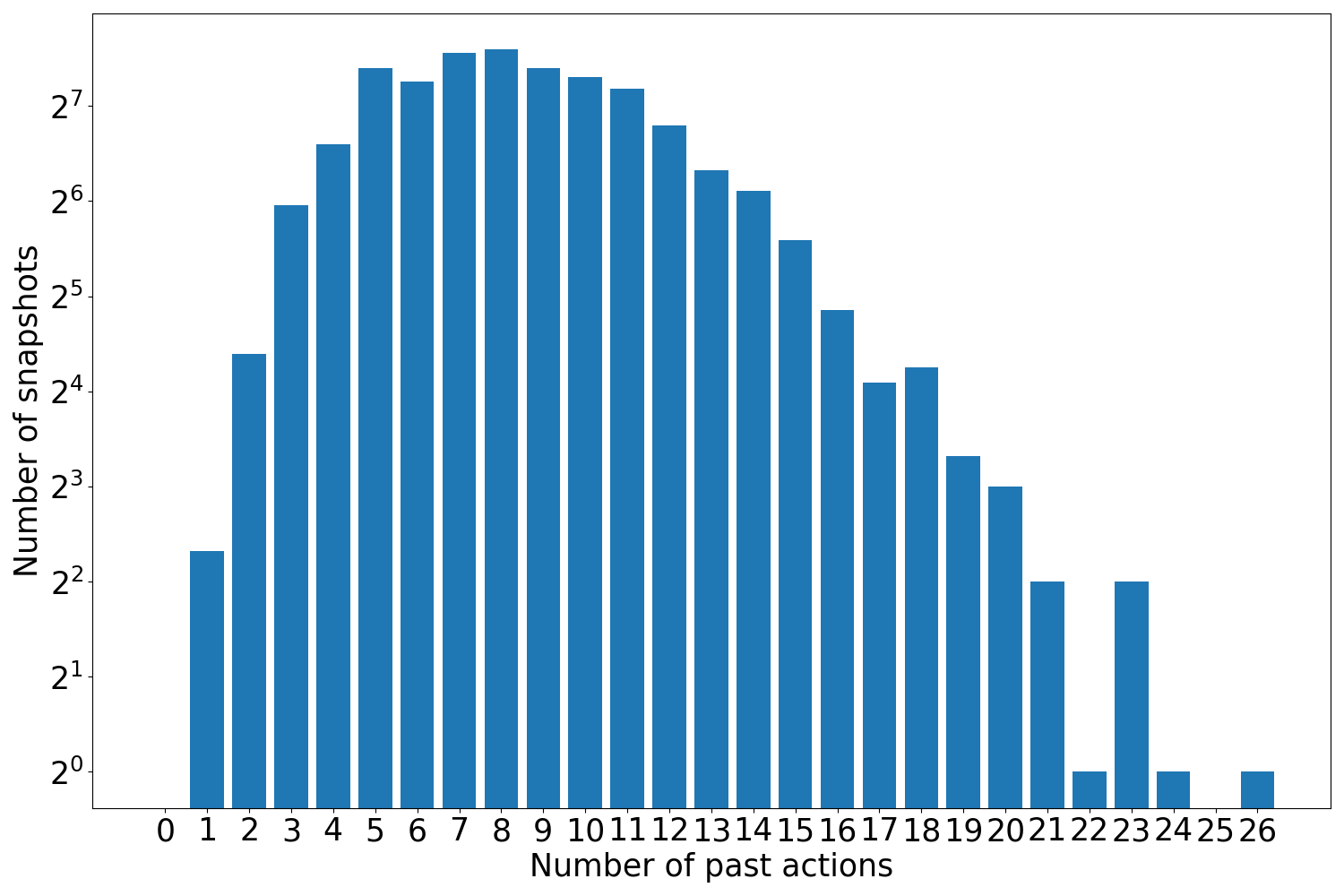}
    \caption{Number of snapshots (in $log_{2}$) for sets of $n$ past actions in the Action Genome test set. (a) -- \textit{Abduct at $T$}, (b) -- \textit{Abduct last snapshot}}.
    \label{fig:freq_dist_setups}
\end{figure}

\noindent
\textbf{How to generate action sets and sequences?}
To obtain the ground truth action set $\mathcal{A}$ for an image in the Action Genome dataset using the Charades action labels, we first compute the time $t$ for each individual frame within a video sequence by using the formula: ${t} = \frac{v_{d}}{n}$,
where $v_{d}$ and $n$ denote the video duration and the number of frames in the video respectively. Then, we multiply the current frame number $f_{n}$ with ${t}$ to obtain the current time,  $t_{c} = t \times f_{n}$.

\noindent\textbf{Action sets:} As each video contains multiple actions, we check whether the current time of the frame $t_{c}$, falls within the start $t_{s}$ and end $t_{e}$ time of the action. If it does, we add the ground truth action label to the action set $\mathcal{A}_{n}$ for the image. To obtain the ground truth action set for the $t^{th}$ image, we combine all previous action sets from $t=1$ up to and including the $t^{th}$ image to form the set.

\noindent\textbf{Action sequences:} We sort the start time $t_{s}$ of the actions contained in the video in ascending order. Then, for each image, if the current time of the frame is greater than the start time of the action ($t_{c} \geq t_{s}$), we add it to the sequence.

We provide details on the number of images for a set of $n$ past actions in the AG dataset for these setups in Figure~\ref{fig:freq_dist_setups}.
As can be seen from these statistics, the majority of the images have more than five actions and some images have as many as 26 actions.

\section{Implementation Details}
We use FasterRCNN \cite{ren2015faster} with a ResNet-101 \cite{he2016deep} backbone to extract human and object features from each image based on the ground truth person and object bounding boxes provided by AG for all object-relational models. 
We load pre-trained weights provided by \cite{cong2021spatial} that were trained on the training set of AG which obtained 24.6 mAP at 0.5 IoU with COCO metrics. 
The parameters of the FasterRCNN during training and inference are fixed for the abductive past action inference task.  
Our default human and object visual representations have 512 dimensions obtained from 2048 dimensional visual features from the FasterRCNN. We use linear mappings to do this. 
During training, we train the models for 10 epochs and set the batch size to 1 video (there are many frames in a video).  {We assume the frames are i.i.d. Note that even though there are multiple images in a batch, the images are processed in parallel and individually for the transformer and graph models respectively. There is no sharing of information between images.} We use the AdamW \cite{loshchilov2017decoupled} optimizer with an initial learning rate of 1e-5 along with a scheduler to decrease the learning rate by a factor of 0.5 to a minimum of 1e-7. We utilize Glove \cite{pennington2014glove} word embedding of size 200 for the human and object semantic features. In addition, gradient clipping with a maximal norm of 5 is applied. Moreover, we report the mean across 3 different runs for each configuration to ensure we report the most accurate performance of our models. All models (except end-to-end \blue{and ViT}) are trained on a single RTX3090 or A5000 GPU. For CLIP, we use publicly available implementations~\cite{radford2021learning}. We use the public API of OpenAI for GPT 3.5 models. 

\section{Additional Experiments}

\subsection{Abductive Past Action Sequence Prediction}
\label{seq.exp.seq}
Next, we formulated the abductive past action sequence prediction task based on the \textit{Abduct at T} setup. 
We attached a GRU / transformer decoder to our existing object-relational models. 
To train both sequence prediction models, we freeze the object detector and relational model ($\phi()$). 
Then, we use the relational vector $x_r$ and action distribution obtained from $\phi_{c}()$ as the initial hidden state and pass it to the GRU respectively. 
The transformer decoder takes non-pooled relational features (a matrix of size $n\times d$) as the key, value, and max-pooled relational features $x_r$ as the query. 
The output of these models is fed into a linear classifier to produce action sequences autoregressively. 
The results of these models are reported in Table \ref{tab:main_results_3}. The BiGED model obtains slightly better performance than the rest. Although the performances of these models are suboptimal, we note that humans are also unable to obtain satisfactory results (only 14.00\% accuracy). As we are constrained to only utilize available information in a single frame, the solution contains a substantial amount of sequence permutations. Therefore, the task is extremely challenging.
The poor human performance also suggests how humans may use abduction. Perhaps humans do not resolve causal chains when performing abduction as it is a very challenging task.

We use the Hamming Loss to evaluate the action sequence prediction models as follows:
\begin{equation}
    H = \frac{1}{N*L}\sum_{n=1}^{N} \sum_{l=1}^{L}\left[ y_{l} \neq \hat{y}_{l} \right]
\end{equation}
where $N$ is the total number of samples and $L$ is the sequence length. Finally, for a given sample, the accuracy is $(1 - H)\times100$.

\begin{table}[t]
\centering
\caption{Abductive past action sequence prediction using the proposed methods on the \textit{Abduct at $T$} setup.}
\resizebox{0.95\columnwidth}{!}{
\begin{tabular}{lcc}
\hline
\textbf{Model} & \multicolumn{2}{c}{\textbf{Accuracy}} \\ \hline
Human performance & \multicolumn{2}{c}{14.00} \\ \hline
\textbf{} & \textbf{GRU} & \textbf{Transformer} \\ \hline
Relational MLP & 9.43\blue{$\pm$0.13} & 9.59\blue{$\pm$0.06} \\
Relational Self Att. Transformer & 9.72\blue{$\pm$0.06} & 9.95\blue{$\pm$0.07} \\
Relational Cross Att. Transformer & 9.69\blue{$\pm$0.18} & 9.96\blue{$\pm$0.12} \\
Relational GNNED & 9.81\blue{$\pm$0.05} & 10.11\blue{$\pm$0.19} \\
RBP & 10.48\blue{$\pm$0.05} & 10.22\blue{$\pm$0.12} \\
BiGED & \textbf{10.54\blue{$\pm$0.15}} &\textbf{ 10.1\blue{$\pm$0.14}} \\ \hline
\end{tabular}
}
\label{tab:main_results_3}
\end{table}


\subsection{Ablation Study}

\textbf{Ablation on graph affinity function:}
By default, we use the Jaccard Vector Similarity as the affinity $W_A(i,j)$ for the GNNED and BiGED models. Here, we ablate the impact of this design choice by comparing it with cosine similarity and dot product. As can be seen from the results in Table~\ref{tab:affinity}, the Jaccard Vector Similarity (JVS) obtains better results than cosine similarity and dot product. This behavior can be attributed to the fully differentiable and bounded nature of JVS in contrast to the dot product or cosine similarity.
\begin{table}[t]
\caption{Ablation on graph affinity using \textit{Abduct at T} setup.}
\centering
\begin{tabular}{lccccc}
\hline
\textbf{Model} & \multicolumn{1}{l}{\textbf{mAP}} &  \textbf{R@10} & \textbf{mR@10} \\ \hline
Jaccard Vector Similarity                   & \textbf{35.75} & \textbf{60.55}  & \textbf{44.37} \\
Cosine Similarity       &  34.17 & 57.98 & 41.97  \\
Dot product             &  28.81  & 54.68 & 38.38 \\
 \hline
\end{tabular}
\label{tab:affinity}
\end{table}

\noindent
\textbf{Impact of semantic features and learning scheduler:} 
Apart from the two different setups mentioned, we also use a third setup for ablations. In the third setup, the action sets are formed from the current and previous images which form the ground truth denoted by $\mathcal{A} = \{\mathcal{A}_{t-1} \bigcup \mathcal{A}_{t} \}$ for faster experimentation. 
We retrain all object-relational models with the corresponding past action set {obtained from the current and previous images.} 
We perform ablation studies on the relational self-attention transformer based on this setup. 
These findings can also be generalized to the other setups as mentioned earlier.

We evaluate the effect of visual and semantic (Glove~\cite{pennington2014glove}) features in Table~\ref{tab:decompose_models}.  
The use of semantic features provides a huge performance boost {across all metrics}. 
We attribute the performance increase to the contextual information provided by the semantics. 
The semantics of objects enable the model to effectively identify and relate actions, providing a more intuitive means for reasoning about these actions.
It is also interesting to see the impact of the learning rate scheduler which provides considerable improvement for the transformer model. Therefore, we use semantics and the learning rate scheduler for all our models. 

\begin{table}[t]
\caption{Ablation study for the impact of semantic features and scheduler on the abductive past action set inference for the \textit{Abduct from current and previous images} setup using self-attention transformer.}
\centering
\resizebox{\columnwidth}{!}{\begin{tabular}{lccccc}
\hline
\textbf{Model} & \multicolumn{1}{l}{\textbf{mAP}} &  \textbf{R@10} & \textbf{mR@10} \\ \hline
Visual only                   & 21.42\blue{$\pm$0.13} & 46.44\blue{$\pm$0.12}  & 34.24\blue{$\pm$0.42} \\
Visual + scheduler            & 21.93\blue{$\pm$0.16}   & 47.04\blue{$\pm$0.44} & 34.80\blue{$\pm$0.47} \\
Visual + semantic             & 35.40\blue{$\pm$0.16}  & 68.47\blue{$\pm$0.06} & 54.90\blue{$\pm$0.52} \\
Visual + semantic + scheduler & \textbf{35.77\blue{$\pm$0.30}} & \textbf{69.16\blue{$\pm$0.50}} & \textbf{55.70\blue{$\pm$0.47}} \\ \hline
\end{tabular}}
\label{tab:decompose_models}
\end{table}

\subsection{Object-Relational Model Parameters}
\begin{table}[t]
\caption{The object-relational model parameters for the abductive past action inference task.}
\centering
\begin{tabular}{lr}
\hline
\multicolumn{1}{c}{\textbf{Model}} & \textbf{Parameters} \\ \hline
Relational MLP                                & 13.4M  \\
Relational Self Att. Transformer                     & 101.2M \\
Relational Cross Att. Transformer                   & 65.9M \\
Relational GNNED                              & 80.7M  \\
RBP                                & 373.4M \\
BiGED                              & 213.6M \\ \hline
\end{tabular}
\label{tab:parameters}
\end{table}
The proposed object-relational model parameters are shown in Table~\ref{tab:parameters}. The rule-based inference model does not have any parameters and is therefore omitted from the table. Based on the results shown earlier, we note that the GNNED model obtains better performance than the transformer model even though it has lesser parameters. In addition, our proposed BiGED model has lesser parameters and performs comparable to or better than the RBP model. These further demonstrate the effectiveness of the proposed GNNED, RBP, and BiGED models for the challenging task of abductive past action inference. 

\begin{table*}[t]
\centering
\resizebox{0.95\textwidth}{!}
{
\begin{tabular}{cp{0.20\textwidth}p{0.20\textwidth}p{0.20\textwidth}p{0.20\textwidth}p{0.20\textwidth}p{0.20\textwidth}p{0.20\textwidth}}
\textbf{Snapshot} & \multicolumn{1}{c}{\textbf{\blue{GT}}} & \multicolumn{1}{c}{\textbf{\blue{Rule-based}}} & \multicolumn{1}{c}{\textbf{MLP}} & \multicolumn{1}{c}{\textbf{Transformer}} & \multicolumn{1}{c}{\textbf{GNNED}} & \multicolumn{1}{c}{\textbf{RBP}} & \multicolumn{1}{c}{\textbf{BiGED}} \\ \hline
    \includegraphics[scale=0.30, valign=T]{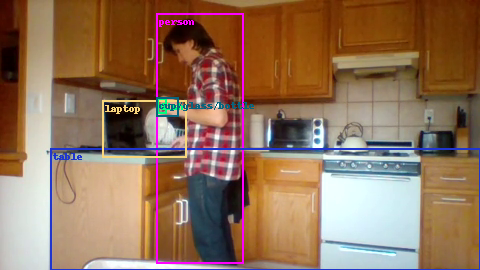} &
    \vspace{-13pt}
    \flushleft
      {Holding a dish} \\
      {Taking a cup/glass/bottle from somewhere} \\
      {Holding a cup/glass/bottle of something} \\
      {Working/Playing on a laptop}\\
      {Working at a table} \\
      {Watching a laptop or something on a laptop} \\
      {Drinking from a cup/glass/bottle}\\
    &
    \vspace{-13pt}
    \flushleft
        \textcolor{red}{Putting something on a table} \\
        \textcolor{green}{Taking a cup/glass/bottle from somewhere} \\
        \textcolor{red}{Sitting at a table} \\
        \textcolor{green}{Working at a table} \\
        \textcolor{green}{Working/Playing on a laptop}\\
        \textcolor{red}{Holding a laptop} \\
        \textcolor{red}{Opening a laptop} \\
        \textcolor{red}{Sitting in a chair} \\
        \textcolor{green}{Drinking from a cup/glass/bottle}\\
        \textcolor{red}{Taking a dish/es from somewhere} \\
      
    &
    \vspace{-13pt}
    \flushleft
        \textcolor{green}{Holding a dish} \\
        \textcolor{red}{Putting something on a shelf} \\
        \textcolor{red}{Walking through a doorway} \\
        \textcolor{red}{Closing a closet/cabinet} \\
        \textcolor{red}{Opening a closet/cabinet} \\
        \textcolor{red}{Putting a dish/es somewhere} \\
        \textcolor{red}{Taking a dish/es from somewhere} \\
        \textcolor{red}{Someone is smiling} \\
        \textcolor{red}{Someone is sneezing} \\
        \textcolor{red}{Someone is standing up from somewhere} \\
    & 
    \vspace{-13pt}
    \flushleft
        \textcolor{green}{Holding a dish} \\
        \textcolor{green}{Taking a cup/glass/bottle from somewhere} \\
        \textcolor{red}{Putting something on a table} \\
        \textcolor{red}{Closing a closet/cabinet} \\
        \textcolor{red}{Opening a closet/cabinet} \\
        \textcolor{red}{Putting a dish/es somewhere} \\
        \textcolor{red}{Taking a dish/es from somewhere} \\
        \textcolor{red}{Someone is cooking something} \\
        \textcolor{red}{Someone is smiling} \\
        \textcolor{red}{Someone is standing up from somewhere} \\

     &  
    \vspace{-13pt}
    \flushleft
        \textcolor{green}{Holding a dish}\\
        \textcolor{green}{Taking a cup/glass/bottle from somewhere}\\
        \textcolor{red}{Putting something on a table}\\
        \textcolor{green}{Working/Playing on a laptop}\\
        \textcolor{red}{Tidying up a table} \\
        \textcolor{red}{Putting a cup/glass/bottle somewhere}\\
        \textcolor{green}{Drinking from a cup/glass/bottle}\\
        \textcolor{red}{Putting a dish/es somewhere} \\ 
        \textcolor{red}{Taking a dish/es from somewhere} \\ 
        \textcolor{red}{Someone is standing up from somewhere} \\
     & 
    \vspace{-13pt}
    \flushleft
        \textcolor{green}{Holding a dish}\\
        \textcolor{green}{Taking a cup/glass/bottle from somewhere}\\
        \textcolor{green}{Holding a cup/glass/bottle of something}\\
        \textcolor{green}{Working/Playing on a laptop}\\
        \textcolor{red}{Putting something on a table} \\
        \textcolor{green}{Watching a laptop or something on a laptop}\\
        \textcolor{green}{Drinking from a cup/glass/bottle}\\
        \textcolor{red}{Putting a cup/glass/bottle somewhere} \\
        \textcolor{red}{Opening a closet/cabinet} \\
        \textcolor{red}{Taking a dish/es from somewhere} \\
     & 
   {\vspace{-13pt}
    \flushleft
        \textcolor{green}{Holding a dish}\\
        \textcolor{green}{Taking a cup/glass/bottle from somewhere}\\
        \textcolor{green}{Holding a cup/glass/bottle of something}\\
        \textcolor{green}{Working/Playing on a laptop}\\
        \textcolor{red}{Putting something on a table} \\
        \textcolor{red}{Walking through a doorway} \\
        \textcolor{red}{Putting a cup/glass/bottle somewhere} \\
        \textcolor{red}{Putting a dish/es somewhere} \\
        \textcolor{red}{Taking a dish/es from somewhere} \\
        \textcolor{red}{Someone is standing up from somewhere}} \\

    \includegraphics[scale=0.30, valign=T]{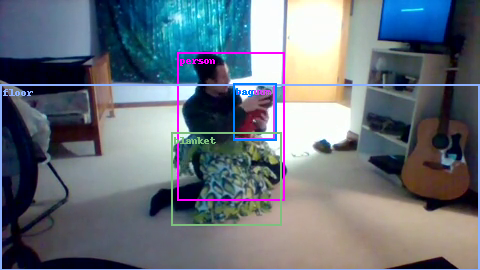} & 
    \vspace{-10pt}
    \flushleft
          {Holding a blanket}\\
          {Holding a bag}\\
          {Snuggling with a blanket}\\
          {Sitting on the floor}\\
          {Lying on the floor}\\
          {Holding a vacuum}\\
          {Someone is awakening somewhere}\\
    &
    \vspace{-10pt}
    \flushleft
        \textcolor{red}{Putting a broom somewhere} \\
        \textcolor{red}{Taking a broom from somewhere} \\
        \textcolor{red}{Throwing a broom somewhere} \\
        \textcolor{red}{Tidying up with a broom} \\
        \textcolor{red}{Fixing a light} \\
        \textcolor{red}{Turning on a light} \\
        \textcolor{red}{Turning off a light} \\
        \textcolor{red}{Drinking from a cup/glass/bottle} \\ 
        \textcolor{red}{Holding a cup/glass/bottle of something} \\
        \textcolor{red}{Pouring something into a cup/glass/bottle} \\
    &
    \vspace{-10pt}
    \flushleft
        \textcolor{green}{Holding a blanket}\\
        \textcolor{red}{Holding some clothes} \\
        \textcolor{red}{Putting clothes somewhere} \\
        \textcolor{red}{Holding a towel/s} \\
        \textcolor{red}{Putting a blanket somewhere} \\
        \textcolor{red}{Taking a blanket from somewhere} \\
        \textcolor{red}{Walking through a doorway} \\
        \textcolor{red}{Someone is smiling} \\
        \textcolor{red}{Someone is sneezing} \\ 
        \textcolor{red}{Someone is standing up from somewhere} \\
     &  
    \vspace{-10pt}
    \flushleft
        \textcolor{green}{Holding a blanket}\\
        \textcolor{green}{Holding a bag}\\
        \textcolor{red}{Holding some clothes} \\
        \textcolor{red}{Putting clothes somewhere} \\
        \textcolor{red}{Taking some clothes from somewhere} \\
        \textcolor{red}{Opening a bag} \\
        \textcolor{red}{Taking a bag from somewhere} \\
        \textcolor{red}{Walking through a doorway} \\
        \textcolor{red}{Someone is smiling} \\
        \textcolor{red}{Someone is standing up from somewhere} \\
     &
    \vspace{-10pt}
    \flushleft
        \textcolor{green}{Holding a blanket}\\
        \textcolor{green}{Holding a bag}\\
        \textcolor{green}{Snuggling with a blanket}\\
        \textcolor{red}{Opening a bag}\\
        \textcolor{red}{Putting a bag somewhere}\\
        \textcolor{red}{Taking a bag from somewhere}\\
        \textcolor{red}{Putting a blanket somewhere}\\
        \textcolor{red}{Taking a blanket from somewhere}\\
        \textcolor{red}{Tidying up a blanket/s}\\
        \textcolor{red}{Someone is standing up from somewhere}\\
     &
    \vspace{-10pt}
    \flushleft
        \textcolor{green}{Holding a blanket}\\
        \textcolor{green}{Holding a bag}\\
        \textcolor{green}{Snuggling with a blanket}\\
        \textcolor{green}{Sitting on the floor}\\
        \textcolor{green}{Holding a vacuum} \\
        \textcolor{red}{Opening a bag} \\
        \textcolor{red}{Taking a bag from somewhere} \\
        \textcolor{red}{Taking a blanket from somewhere}\\
        \textcolor{red}{Drinking from a cup/glass/bottle} \\
        \textcolor{red}{Tidying something on the floor} \\
     &  
    {\vspace{-10pt}
    \flushleft
        \textcolor{green}{Holding a blanket}\\
        \textcolor{green}{Holding a bag}\\
        \textcolor{green}{Snuggling with a blanket}\\
        \textcolor{green}{Sitting on the floor}\\
        \textcolor{green}{Holding a vacuum} \\
        \textcolor{red}{Opening a bag} \\
        \textcolor{red}{Taking a bag from somewhere}\\
        \textcolor{red}{Throwing a blanket somewhere}\\
        \textcolor{red}{Fixing a vacuum}\\
       \textcolor{red}{Someone is standing up from somewhere}}\\
    \end{tabular}}
    
    \captionof{figure}{Manually selected qualitative results produced by each model on the abductive past action set inference: \emph{Abduct last image} setup on the AG test dataset. The first column shows the image \blue{followed by their corresponding ground truth past actions.}
    \blue{The remaining columns display the actions predicted by each model, with correct predictions highlighted in green and incorrect predictions highlighted in red.}}
\label{fig:qualitative-results}
\end{table*}

\subsection{Qualitative Results}
We compare qualitative results for the abductive past action set prediction task in Figure \ref{fig:qualitative-results}. Depending on the number of past action labels an image has, we take the same number of top-k predicted actions from each model. All models demonstrate their ability to perform abductive past action inference. 
In the first image, there are objects such as a person, laptop, table, cup, and dish. In the second image, there are objects such as a person, floor, blanket, bag, and vacuum. 
In both scenarios, RBP and BiGED demonstrate that they can infer past actions more accurately.

\end{document}